# Semantic Polarity of Adjectival Predicates in Online Reviews

**Ae-Lim Ahn\*, Eric Laporte\*\* and Jee-Sun Nam\***

\*DICORA, Hankuk University of Foreign Studies, Korea/ \*\*LIGM, Université Paris-Est, France

polarbeariii@hotmail.com, Eric.Laporte@univ-paris-est.fr, namjs@hufs.ac.kr

**Abstract**

Web users produce more and more documents expressing opinions. Because these have become important resources for customers and manufacturers, many have focused on them. Opinions are often expressed through adjectives with positive or negative semantic values. In extracting information from users' opinion in online reviews, exact recognition of the semantic polarity of adjectives is one of the most important requirements. Since adjectives have different semantic orientations according to contexts, it is not satisfying to extract opinion information without considering the semantic and lexical relations between the adjectives and the feature nouns appropriate to a given domain. In this paper, we present a classification of adjectives by polarity, and we analyze adjectives that are undetermined in the absence of contexts. Our research should be useful for accurately predicting semantic orientations of opinion sentences, and should be taken into account before relying on an automatic methods.

## 1. Introduction

With the high increase in the number of documents expressing opinions, Web opinion mining is becoming a challenging task. Users express their opinions about products on the web, and people share their opinions. These opinions become important resources for customers who want information about products and manufacturers who wish to improve their productivity. Therefore, the demand for automatic extraction of opinions from Web documents is increasing, and the research on classification of reviewers' opinion progresses steadily. Such studies are called Opinion Mining(OM), which covers a range of activities from retrieving opinion sentences in web documents to determining their meaning. Opinion mining in Korean web documents resorts to increasingly various approaches. However, serious linguistic analyses about opinion documents are still rare. In our opinion, in order to get efficient results from opinion mining, fundamental work on opinion sentences and construction of linguistic resources have to be performed in advance.

A sentence which contains one or more topic segments denoting product features, and one or more evaluative segments expressing opinions, is called an "opinion sentence" (Hu 2004). Such a sentence contains opinion words. The majority of opinion words are adjectival predicates. Therefore, to deduce the orientation of an opinion sentence, we examine semantic the polarity of adjectival predicates (i.e. the positive value [+] vs. the negative value [-]). However, some adjectives are context-dependent. That is, a given word may indicate different opinions depending on its domain, or even within one and the same domain, depending on product features. For example, "The battery life is *long*." expresses a positive opinion (+); and "It takes a *long* time to focus." a negative opinion (-), with the same opinion word "long" combined with distinct product features in the same domain (Ding and Liu 2007). Many OM researchers point out the ambiguity of adjectives, and emphasize the necessity of analysing them. Korean sentences are little different from English of French sentences: Korean adjectives may have context dependent polarity. We should consider semantic and lexical restrictions between adjectives and the co-occurring features (or topics) in one domain.

In addition, adjectives can mark serve as intensity markers for other opinion words. For example, in *"I yenghwa-nun kwankayk-ul ppalatuli-nun hupiplyek-I **kang**-haysseyo."*("This movie has **strong** attraction for audiences."), *"kanghata"*("strong") reinforces *"hupiplyek"*("attraction"), and itpresents a positive opinion. In other hands, in *"Phoklyekseng-I **kang**-han yenghwa-tukunyo."*( "It was a movie of **strong** violence."), *"kanghata"* does the same for *"Phoklyekseng"*("violence") which presents a negative opinion. As we can see, opinion mining should not be limited to simply counting "good" and "bad" words in a document. In this paper, we define the features that people consider when they evaluate the products and determine the polarity of adjectives depending on their features within various domains.

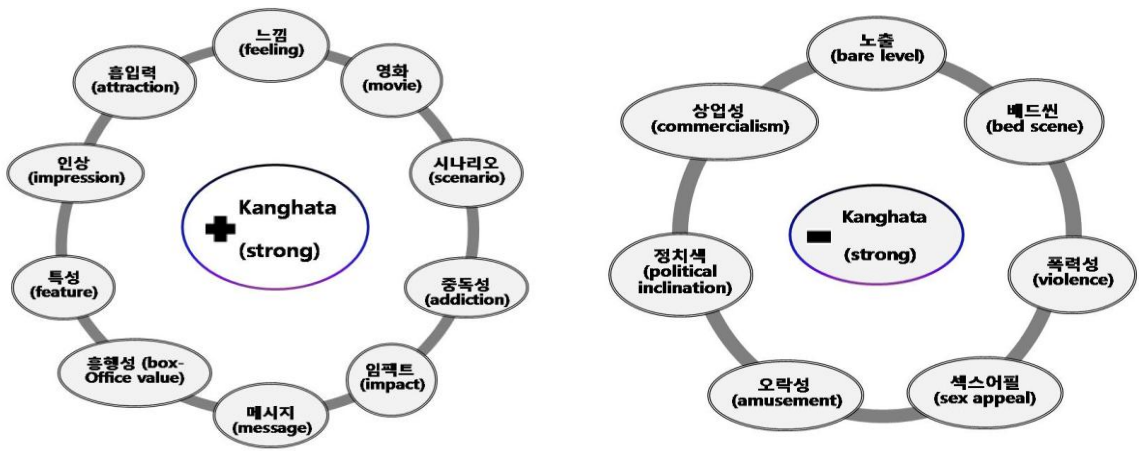

**Figure 1.** The polarity of *"Kanghata"* ("strong") depending on features in the MOVIE domain.

Section 2 will survey previous research related to opinion mining. We classify adjectival predicates by semantic polarity in Section 3, and we present feature lists in each domain in Section 4. Then a case study on a specific adjective is reported in Section 5. Section 6 includes the conclusion and suggests future studies[1].

[1] This work was supported by the French Unique Interministerial Fund (FUI) and by French industrial cluster Cap Digital, through the funding of the DoXa project on automatic processing of opinion and sentiment.

## 2. Related Work

Studies on the extraction of opinion sentences show that adjectives often express users' opinion. Classification of opinions is conducted at document-level or sentence-level in recent studies. That is, the document can be interpreted as positive or negative as a whole, or each sentence is classified as positive or negative (Yuk 2008). Most studies adopt machine learning or fully automated approaches to assign a polarity to documents or sentences. One method for labeling positively and negatively oriented adjectives uses conjunctions. It is an unsupervised learning method for obtaining polarity of adjectives with accuracy over 90% (Hatzivassiloglou & Mckeown 1997). Extraction using seed words extends an initial set of words with predetermined orientation labels to construct a larger set of semantically oriented words (Turney 2002). . This method finds that some adjectives have neutral orientation, because it does not consider semantic ambiguity depending on domains. As an attempt to solve this limitation, correlations between the seed words and other adjectives are computed so as to enrich the sets of seed words with new domain-relevant opinion adjectives (Harb et al 2008). This approach still produces much noise in certain types of text. Without linguistic analysis of each adjective, this problem is difficult indeed.

For Korean sentiment classification, several natural language processing techniques, including the use of a semantic dictionary, have been used. The semantic dictionary contains words used to express product features and customer's opinions: it was constructed semi-automatically (Myeng et al. 2008). This dictionary was extended into a detailed description of opinion features, opinion words, independent opinion words, opinion phrases, and some negation expressions (Yuk 2008).

In France, two partners of the DoXa project[2], Arisem and the LIGM laboratory, have undertaken the construction of language resources for a few domains, describing the following data: the vocabulary of opinions and sentiments (Brizard, 2009; Varga, 2009); the words and expressions used to refer to topics and subtopics in evaluative documents, such as "*durée de vie de la batterie*"("battery life"), these resources being structured as ontologies in the OWL-DL language; the markers used to intensify or attenuate the expression of opinions and sentiments, e.g. *fort* "strong".

In this paper, we describe a linguistic analysis of polarity of adjectives by considering various contexts to improve the accuracy of automatic systems. The goal of this study will be the construction of an "*Opinion-Feature Dictionary*" based on this analysis.

## 3. Classification of Adjectives by Semantic Polarity

Opinion classification is based on the meaning of words and their relations. As semantic orientation of opinion sentences is most affected by the polarity of the adjectives, study on polarity of adjectives is required. By analyzing a corpus built from Web documents, we can classify the observed adjectives

---

[2] DoXa is a French project on automatic processing of opinion and sentiment. https://www.projet-doxa.fr/

according to their semantic polarity.

|  | Cosmetic products | Hotel | Hospital | Mobile Phone | Movie |
|---|---|---|---|---|---|
| **Web site** | www.beauu.com | www.hetelnjoy.com | www.ysps.com | www.cetizen.com | www.moviejoy.com www.maxmovie.com |
| **Size** | 23,502 tokens | 23,505 tokens | 26,854 tokens | 23,776 tokens | 24,631 tokens |
| **Adjectival predicates** | 321 types | 278 types | 297 types | 235 types | 291 types |

**Table 1. Five domains size for our corpora**

We classified adjectives using five domains represented by five corpora of online texts: movies [MOVIE], cosmetic products [COSMETIC PRODUCT], hospitals [HOSPITAL], mobile phones [MOBILE] and hotels [HOTEL]. These corpora consist of about 23,500 ~ 26,800 tokens each.

We obtained the list of the most frequent adjectives in the five corpora, using the Korean lexical analyzer *Geuljabi*[3]. The frequent adjectival predicates are different depending on their domains (Table 1). This means the features which people consider most significant for evaluation are different in each domain. For example, reviews about mobile phones contain adjectival predicates which express an evaluation of the functions or look of mobile phones, such as *"ppaluta"*("fast") and *"mwukupta"*("heavy"). In reviews about movies, emotional adjectives occur with high frequency, such as *"caymiissta"*("interesting") and *"sulputa"*("sad"). Consequentially, the choice of adjectives differs for each domain. This explains why adjectives can have different polarity depending on the features they express an opinion about.

We assigned a semantic polarity to adjectives extracted from the corpora, whose types are about 300. We classified adjectival predicates into two types: adjectives of ABSOLUTE polarity and those of RELATIVE polarity.

**A. Type 1: Adjectives of ABSOLUTE Polarity**
    A-1. Positively-Oriented Adjectives
      Examples: *cohta*("good"), *yeypputa*("beautiful"), *phyenhata*("comfortable"), *olhta*("correct"), *chinhata*("friendly") etc.
    A-2. Negatively-Oriented Adjectives
      Examples: *silhta*("unpleasant"), *simhata*("harsh"), *akkapta*("wasteful"), *telepta*("dirty"), *sikkulepta* ("noisy") etc.

**B. Type 2: Adjectives of RELATIVE Polarity**
    Examples: *kapyepta*("light"), *mukepta*("heavy"), *kanghata*("strong"), *kiphta*("deep"), *khuta*("big"), *nophta*("high") etc.

---
[3] For more detailed information, see "www.sejong.or.kr"

Adjectives of ABSOLUTE polarity do not depend on contexts. Their polarity is stable across domains and features. On the other hand, adjectives of RELATIVE polarity have context-dependent polarity. Their polarity is affected by several factors in their sentences.

Based on this classification, we obtain the frequency of each type in the five domains (Table 2). The average of the total number of occurrences of adjectives is 2,135 per domain: 884 occurrences of adjectives of ABSOLUTE polarity and 1,251 of adjectives of RELATIVE polarity.

|  | Cosmetic Product | Hotel | Hospital | Mobile Phone | Movie |  | Average |
|---|---|---|---|---|---|---|---|
| **Total frequency of adjectival predicates** | 2,151 | 2,262 | 2,174 | 1,901 | 2,186 |  | 2,135 |
| **ABSOLUTE polarity** | 912 | 766 | 1,022 | 814 | 908 |  | 884(41.4%) |
| **RELATIVE polarity** | 1,239 | 1,496 | 1,152 | 1,087 | 1,278 |  | 1,251(58.6%) |

**Table 2. The frequency of adjectives bearing ABSOLUTE or RELATIVE polarity in various corpora**

Adjectives of RELATIVE polarity account for more occurrences than adjectives of ABSOLUTE polarity. This shows that the current keyword-based approach in which adjectives bear fixed polarities has a serious drawback. If we perform keyword-based extraction method with adjectives of RELATIVE polarity, we cannot obtain reliable results. For instance, in the HOTEL and MOBILE corpora, when we extract sentences containing *"khuta"*("big"), we will get about 20% noise.

|  | **HOTEL** | **MOBILE** |
|---|---|---|
| **Total concordance** | 44 sentences | 53 sentences |
| **Noise** | 9 sentences | 12 sentences |
| **Total Opinion Sentence** | 35 sentences | 41 sentences |

**Table 3. The result of sentence extraction by *Khuta***

Table 3 shows that keyword-based extraction allows only about 79% and 77% precision respectively. The Noise row shows the number of Non-opinion sentences, which express facts in a neutral way. Let us compare (1a) to (1b).

(1) a. *"Lostey hotheyl cupyen-ey **khun** kenmul-i manh-supnita."*
   (" There are ***big*** buildings around the hotel.")

   b. *"Hotheyl kyumo-ka **khu**-ko kunsaha-neyyo."*
   ("The hotel is ***big*** and wonderful.")

Both sentences in (1) have *"khuta"* in the predicate. However, sentence (1a) expresses a fact; sentence (1b) expresses an opinion about hotel. This means that not every sentence with *"khuta"* is an opinion. Some previous research uses an annotated corpus, where opinions and facts are tagged, but it is not possible to obtain reliable annotated corpora in sufficient quantity. In addition, determining whether adjectives express an opinion or a fact is not done in perfect way. Therefore, we require close observation of sentences classified by the polarity of their adjectives. Especially, sentences containing adjectives of RELATIVE polarity have to be analyzed closely. We examine them in Section 5.

**3.1. Adjectives of Absolute Polarity**

Adjectives of ABSOLUTE polarity are not affected by contexts and can be interpreted independently of domains. We determine their polarity to construct the Lexicon of ABSOLUTE Polarity. Frequent adjectives can be different depending on domains, but their semantic polarity does not change. Hence, we need not present information about domains(or features) in this lexicon.

| *Adjective* | *Polarity* | *Adjective* | *Polarity* | *Adjective* | *Polarity* |
|---|---|---|---|---|---|
| 가능하다(possible) | + | 밉다(hateful) | - | 예민하다(sensitive) | - |
| 간편하다(convenient) | + | 부족하다(insufficient) | - | 예쁘다(pretty) | + |
| 갑갑하다(stuffy) | - | 부지런하다(diligent) | + | 옳다(correct) | + |
| 걱정스럽다(worried) | - | 불행하다(unhappy) | - | 완벽하다(perfect) | + |
| 고맙다(grateful) | + | 비싸다(expensive) | - | 위험하다(dangerous) | - |
| 굉장하다(fantastic) | + | 사랑스럽다(lovely) | + | 유명하다(famous) | + |
| 귀엽다(cute) | + | 사소하다(trivial) | - | 이상하다(strange) | - |
| 깔끔하다(tidy) | + | 상냥하다(affectionate) | + | 잘생기다(handsome) | + |
| 나쁘다(bad) | - | 새롭다(new) | + | 재미있다(interesting) | + |
| 낡다(used) | - | 서투르다(unskilled) | - | 젊다(youthful) | + |

Table 4. Example of ABSOLUTE polarity of adjectives

**3.2. Adjectives of RELATIVE Polarity**

The meanings of adjectives of RELATIVE polarity are affected by co-occurring nouns, sentence patterns or contexts. Since we cannot determine their polarity out of context, further analysis of them and description through linguistic observation is required.

| *Adjective* | *Polarity* | *Adjective* | *Polarity* | *Adjective* | *Polarity* |
|---|---|---|---|---|---|
| 가깝다(near) | +/- | 단순하다(simple) | +/- | 작다(small) | +/- |
| 가볍다(light) | +/- | 두껍다(thick) | +/- | 잦다(frequent) | +/- |
| 강하다(strong) | +/- | 딱딱하다(hard) | +/- | 적다(little) | +/- |
| 길다(long) | +/- | 많다(many) | +/- | 좁다(narrow) | +/- |
| 깊다(deep) | +/- | 멀다(far) | +/- | 짧다(short) | +/- |
| 낮다(low) | +/- | 무겁다(heavy) | +/- | 느리다(slow) | +/- |
| 넓다(wide) | +/- | 빠르다(quick) | +/- | 어리다(young) | +/- |
| 높다(high) | +/- | 얇다(thin) | +/- | 약하다(weak) | +/- |

Table 5. Example of RELATIVE polarity of adjectives

**4. Construction of feature lists of each domain**

We need to study the possible ways of expressing opinions for a given domain, and we can guess which opinions about products customers or manufacturers want to be aware of. One adjective can qualify a finite number of features within a given domain, so we can describe these relations between topic segments and adjectives. By determining the semantic polarity of adjectival predicates when they are applied to a given feature, we can also determine the semantic polarity of corresponding opinion sentences. Through the analysis of the corpora, we define topic categories which reviewers evaluate about products within each domain. We extract nouns from each corpus using *Geuljabi*, and classify feature nouns depending on topic categories.

**4.1 Cosmetic Products**

What evaluative subjects interest customers and manufactures regarding cosmetics? The reviewers evaluate components of cosmetic products such as color, scent and ingredients. Effects of cosmetic products are a major subject of such evaluations, which mean how effective a product is. Reviewers describe strong or weak points against physical symptoms. Price and design of products are also important considerations.

| Domain | Topic Category | Feature Nouns |
|---|---|---|
| **COSMETIC PRODUCT** | *Component* | 색상("color"), 향("scent"), 화학성분("chemicals"), 알코올("alcohol"), 양("quantity"), 비타민("vitamin"), 수분("moisture"), 촉감("touch") etc. |
| | *Effect* | 효과("effect"), 반응("reaction"), 기능("function"), 발림("application"), 흡수("absorption"), 지속력("resistance"), 차단력("protection") etc. |
| | *Physical Symptom* | 지성("oiliness"), 기미("freckles"), 여드름("pimple"), 상처("scar"), 각질("keratin"), 주름("winkle"), 손상("demage"), 자극("irritant") etc. |
| | *Price* | 가격("price"), 세일("sale") etc. |
| | *Design* | 케이스("case"), 튜브("tube"), 모양("shape"), 크기("size") etc. |

Table 6.  Features for evaluation in COSMETIC PRODUCT

**4.2 Hotels**

For HOTEL reviews, we define six topic categories. Reviewers evaluate hotel facilities and supplies. In addition, they describe how the staff provides services, how clean the hotel is and how good the food is. They also give the value for the location, view from the room, and transportations.

| Domain | Topic Category | Feature Nouns |
|---|---|---|
| **HOTEL** | *Facilities* | 호텔("hotel"), 건물("building"), 로비("lobby"), 방("room"), 창("window"), 주차장("parking lot"), 엘리베이터("elevator") etc. |

|  | *Supplies in hotel* | 침대("bed"), 컴퓨터("computer"), 냉장고("refrigerator"), 욕조("bath"), 샴푸("shampoo"), 비누("soap"), 수건("towel") etc. |
|---|---|---|
|  | *Service* | 예약("reservation"), 체크아웃("check-out"), 체크인("check-in"), 안내("guidance"), 룸서비스("room service"), 서비스("service") etc. |
|  | *Cleanliness* | 청소("cleaning"), 냄새("scent"), 먼지("dust"), 정돈("arrangement"), 관리("care") etc. |
|  | *Food* | 조식("breakfast"), 점심("lunch"), 음식("food"), 메뉴("menu"), 레스토랑("restaurant"), 맛("taste"), 음료("drinks"), 빵("bread") etc. |
|  | *Surroundings* | 위치("location"), 전망("view"), 거리("distance"), 길("way"), 야외("outdoor"), 교통("transportation") etc. |

Table 7. Features for evaluation in HOTEL

**4.3 Hospitals**

Reviews of hospitals are increasingly numerous in specific fields such as Plastic Surgery and Dentistry, because results can be definitely different according to doctor's ability or experience. Facilities and services are considerable subjects as much as the doctor's ability nowadays.

| Domain | Topic Category | Feature Nouns |
|---|---|---|
| **HOSPITAL** | *Facilities* | 병원("hospital"), 건물("building"), 시설("facilities"), 인테리어("interior"), 대기실 ("waiting room") etc. |
|  | *Ability and service of staffs* | 의사("doctor"), 간호사("nurse"), 서비스("service"), 실력("ability"), 코디네이터("coordinator"), 상담가("consultant"), etc. |
|  | *Symptom of body* | 통증("pain"), 멍("bruise"), 상처("wound"), 부작용("side effect"), 부기("swelling"), 주름("wrinkle") etc. |
|  | *Result* | 효과("effect"), 회복("recover"), 결과("result"), 변화("change"), 이미지("image"), 모습("appearance"), 콤플렉스("complex") etc. |
|  | *Price* | 가격("price"), 세일("sale") etc. |
|  | *Time* | 대기시간("waiting"), 회복시간("recovery time"), 수술시간("operation time") etc. |

Table 8. Features for evaluation in PLASTIC SURGERY

**4.4 Mobile Phones**

The IT field covers products such as mobile phones, cameras, and PCs. Reviews of mobile phones are most numerous, because it is considered a necessity of life. When people choose their new phone, they consider the look, various functions, and other qualities of mobile phones. Generally, mobile phones are compared about various features.

| Domain | Topic Category | Feature Nouns |
|---|---|---|
| MOBILE PHONE | *Part of mobile phone* | 카메라("camera"), 화면("screen"), 배터리("battery"), 케이스("case"), 스피커("speaker"), 버튼("button") etc. |
| | *Quality* | 속도("speed"), 시간("duration"), 음질("sound"), 해상도("definition"), 움직임("movement"), 접속("connection"), etc. |
| | *Function* | 문자("text message"), 게임("game"), 전화("call"), 사진("picture"), 벨소리("ring"), 사전("dictionary") etc. |
| | *Price* | 가격("price"), 세일("sale") etc. |
| | *Design* | 색깔("color"), 모양("shape"), 크기("size"), 디자인("design") etc. |

**Table 9.  Features for evaluation in MOBILE PHONE**

### 4.5 Movies

Reviews are not limited to the evaluation of material products, but include sentiments on performances such as movies, concerts and musicals. In this case, the reviewers express their emotions towards the performance. In addition to their sentiments on a whole movie, people tell their opinion about the actors' performance and the contents of the story.

| Domain | Topic Category | Feature Nouns |
|---|---|---|
| MOVIE | *Character and Director* | 감독("director"), 인물("character") 배우("actor"), 주인공("protagonist"), 스타("star"), 역할("role"), 작가("writer"), 제작자("producer") etc. |
| | *Story* | 이야기("story"), 줄거리("plot"), 시리즈("series"), 장면("scene"), 사건("episode"), 결말("ending"), 갈등("trouble"), 구조("structure"), 주제("theme") etc. |
| | *Result* | 흥행("box-office"), 평가("evaluation"), 성공("success"), 인기("popularity"), 반응("reaction"), 실패(failure) etc. |
| | *Elements of movie* | 음악("music"), 대사("line"), 출연("casting"), 표현("expression"), 구성("composition"), 묘사("description"), 배경("background"), 목소리("voice") etc. |
| | *Emotion* | 경험("experience"), 기억("memory"), 관심("interest"), 매력("attraction"), 만족("satisfaction"), 감동("impression"), 걱정("worry") etc. |

**Table 10.  Features for evaluation in MOVIE**

## 5. A case study on the Adjective of RELATIVE Polarity "*Khuta*" ("big")

In this section, we analyze opinion sentences with an adjective of RELATIVE polarity, and show restrictions between the adjective and the co-occurring features. We choose *"khuta"*("big"). It occurs in every domain at a high frequency because it is more ambiguous than other adjectives such as *"ppaluta"*("fast") and *"twukkepta"*("thick"). The same word *"khuta"* expresses size, or qualifies various types of magnitude: *"Hwamyen-I nemu **khu**-yo."*(" The screen is very **large**.") describes size of screen, *"Peylsoli-ka nemu **khup**-nita."*("The ring sound is **loud**.") sound volume, and *"Caphan sayong-e **khu**-n cangcem-i isseyo"*("There is a **big** advantage in using the keypad.") the importance of function. In the contrary, *"ppaluta"* expresses only speed, and *"twukkepta"* the distance between sides.

*"Khuta"* generally expresses a favorable opinion in the HOTEL domain, when it evaluates the size of the hotel as in (2a) below. It expresses an unfavorable opinion in the MOBILE PHONE domain, when it evaluates the size of mobile phone like (2b). Even with the same feature, its polarity may depend on the domain. In addition, in one domain, its polarity may depend on features. When *"khuta"* is applied to parts of a mobile phone, such as screen and buttons, it has a positive value as shown in (2c).

(2) a. *"Lostey hotyel-un **khu**-ko wungcang-haysseyo."*
    (" The LOTTE hotel was **big** and magnificent.")

   b. *"Aiphon-uy khuki-ka sayngkak-pota **khu**-n kes kath-ayo."*
    (" The Iphone is **bigger** than I expected.")

   c. *"Hayntuphon pethun-i **khe**-se cal nullye-yo."*
    ("A **big** button on a phone is easy to press.")

Consequently, prediction of polarity of adjectives cannot be achieved at the document or sentence-level, but only by matching jointly the feature and the adjective. Thus, we should consider other contexts that affect the polarity of opinion sentences. This approach leads us to construct an *"Opinion-Feature Dictionary"*. We show examples of an *"Opinion-Feature Dictionary"* of the MOBILE PHONE domain.

| DOMAIN | MOBILE PHONE | | | | |
|---|---|---|---|---|---|
| CATEGORY | *Part of mobile phone* | | | | |
| | 카메라 ("camera") | 화면 ("screen") | 배터리 ("battery") | 케이스 ("case") | 버튼 ("button") |
| 크다("big") | + | + | - | - | + |
| 많다("abundant") | + | z | + | z | - |

| DOMAIN | MOBILE PHONE | | | | |
|---|---|---|---|---|---|
| CATEGORY | *Quality* | | | | |
| | 속도 ("speed") | 음질 ("sound") | 해상도 ("definition") | 접속 ("connection") | 시간 ("duration") |
| 크다("big") | z | + | + | z | z |
| 많다("abundant") | z | z | + | z | z |

| DOMAIN | MOBILE PHONE | | | | |
|---|---|---|---|---|---|
| CATEGORY | *Function* | | | | |
| | 문자("text message") | 게임 ("game") | 전화 ("call") | 사진 ("picture") | 벨소리 ("ring") |
| 크다("big") | z | z | z | + | + |
| 많다("abundant) | z | + | z | z | z |

| DOMAIN | MOBILE PHONE | | | | |
|---|---|---|---|---|---|
| CATEGORY | *Price and Design* | | | | |
| | 색깔 ("color") | 모양 ("shape") | 크기 ("size") | 디자인 ("design") | 가격 ("price") |
| 크다("big") | z | - | - | z | - |
| 많다("abundant") | + | z | z | + | - |

Table11. Sample form Opinion-Features Dictionary of MOBILE PHONE domain

We mark positive polarity with + and negative polarity with −. The features which are not predicated by the adjectives are marked with "z"(zero). Feature noun lists can be extended with synonyms. For example, *"kakyek"*("price") has synonyms such as *"piyoung"*, *"wenga"*, *"cengga"* and *"kumayk"*.

## 6. Conclusion

In extracting information about users' opinion from online reviews, exact recognition of the semantic polarity of the adjectives is one of the most important requirements. As adjectives have different semantic orientations according to their contexts, opinion information cannot be extracted satisfactorily without considering the semantic and lexical relations between adjectives and appropriate feature nouns.

Research about opinion mining with adjectival predicates is active, but linguistic properties of adjectives are underexploited. In this paper, we suggested a classification of adjectives by semantic polarity, and presented an approach to predicting the orientation of opinion sentences with adjectives. We emphasize the importance of describing adjectives of RELATIVE polarity depending on domains and contexts. The ultimate goal of our research will be to construct an *Opinion-Features Dictionary* classified by Domains, which will be necessary to extract accurately online users' opinions from Web documents.

In future work, we need to consider other cases of ambiguity of adjectives of RELATIVE polarity.

(3) a. *"Kheyisu-eyse yakkan **nolan**bich-i nayo."*
   (" The case has a bit of **yellow**.")

   b. *"Sol-i **tungk**-un phyen-ipnita."*
   ("The brush is **round**.")

   c. *"Hayntuphon-i **thupakha**-ki-potan **tungk**-un phyen-ieyyo."*
   ("The mobile phone is **round** rather than **rough**.")

*"Nolahta"*("yellow") in (3a) and *"tungkulta"*("round") in (3b) describe the color and shape of the product, but they do not convey the reviewer's subjective opinion. These sentences are factual and neutral. However, *"tungkulta"* in (3c) expresses a positive opinion. In this comparative sentence, it is contrasted with *"thupakhata"*("rough") which has a negative meaning. Therefore, when we predict the semantic orientation of opinion sentences, we also need an understanding of such sentence structures.